\documentclass[journal,twoside,web]{ieeecolor}
\usepackage{tmi}
\usepackage{cite}
\usepackage{amsmath,amssymb,amsfonts}
\usepackage{algorithmic,algorithm}
\usepackage{tikz}
\usetikzlibrary{decorations.pathreplacing,calc}
\newcommand{\tikzmark}[1]{\tikz[overlay,remember picture] \node (#1) {};}
\newcommand*{\AddNote}[4]{%
    \begin{tikzpicture}[overlay, remember picture]
        \draw [decoration={brace,amplitude=0.5em},decorate,ultra thick,black]
            ($(#3)!(#1.north)!($(#3)-(0,1)$)$) --  
            ($(#3)!(#2.south)!($(#3)-(0,1)$)$)
                node [align=center, text width=1.5cm, pos=0.5, anchor=west] {#4};
    \end{tikzpicture}
}%
\usepackage{graphicx}
\usepackage{textcomp}
\usepackage{caption}
\usepackage{subcaption}
\usepackage{hyperref}
\usepackage{soul}
\def\BibTeX{{\rm B\kern-.05em{\sc i\kern-.025em b}\kern-.08em
    T\kern-.1667em\lower.7ex\hbox{E}\kern-.125emX}}
\usepackage{bbm}
\usepackage{bm}
\usepackage{nicefrac}
\usepackage{mathtools}
\usepackage{verbatim}
\usepackage{microtype} 
\usepackage{listings}
\usepackage{array}
\usepackage{wrapfig}
\usepackage{textcomp}
\usepackage{url}            
\usepackage{booktabs}       

\newcommand{\vrm}{\mathrm{v}}

\newcommand{\Mbf}{\mathbf{M}}

\newcommand{\Pcal}{\mathcal{P}}

\newcommand{\RR}{\mathbb{R}}
\newcommand{\NN}{\mathbb{N}}

\newcommand{\II}{\textbf{I}}
\newcommand{\CC}{\mathbb{C}}
\newcommand{\EE}{\mathbb{E}}

\newcommand{\F}{\mathcal{F}}

\newcommand{\N}{\mathcal{N}}

\newcommand{\dd}{\mathrm{d}}

\newcommand{\Scal}{\mathcal{S}}

\begin{document}
\title{Diffusion Modeling with Domain-conditioned Prior Guidance for Accelerated MRI and qMRI Reconstruction}
\author{Wanyu Bian, Albert Jang, and Fang Liu
\thanks{This work was supported by the grants R21EB031185; R01AR081344; R01AR079442. }
\thanks{W. Bian, A. Jang and F. Liu are with the
Athinoula A. Martinos Center for Biomedical Imaging, Massachusetts General Hospital
and Harvard Medical School, Charlestown, MA 02129 USA. (corresponding author: F. Liu, email:fliu12@mgh.harvard.edu)}
}

\maketitle

\begin{abstract}
This study introduces a novel approach for image reconstruction based on a diffusion model conditioned on the native data domain. 
Our method is applied to multi-coil MRI and quantitative MRI reconstruction, leveraging the domain-conditioned diffusion model within the frequency and parameter domains. 
The prior MRI physics are used as embeddings in the diffusion model, enforcing data consistency to guide the training and sampling process, characterizing MRI k-space encoding in MRI reconstruction, and leveraging MR signal modeling for qMRI reconstruction. Furthermore, a gradient descent optimization is incorporated into the diffusion steps, enhancing feature learning and improving denoising.
The proposed method demonstrates a significant promise, particularly for reconstructing images at high acceleration factors. Notably, it maintains great reconstruction accuracy and efficiency for static and quantitative MRI reconstruction across diverse anatomical structures. Beyond its immediate applications, this method provides potential generalization capability, making it adaptable to inverse problems across various domains.
\end{abstract}


\section{Introduction}
\label{sec:introduction}
Magnetic resonance imaging (MRI) stands as an indispensable, non-invasive imaging tool, pivotal in both medical diagnosis and clinical research. Though MRI delivers unparalleled diagnostic value, its imaging time is lengthy compared to other imaging modalities, limiting its patient throughput. This limitation has galvanized innovations to accelerate the MRI process, all with the common goal of drastically reducing scan time without compromising image quality\cite{pruessmann1999sense,griswold2002grappa}. Recently, deep learning has shown great potential in addressing this issue. Numerous techniques have been introduced that enhance the performance of optimization algorithms using finely tuned sophisticated neural networks, achieving excellent results\cite{liang2020deep}. A substantial portion of these state of the art methods utilize conditional models, adeptly converting undersampled data inputs into outputs accurately mirroring the fully-sampled data acquisitions \cite{sun2016deep,schlemper2017deep,hammernik2018learning,mardani2018deep,zhu2018image,liu2020deep,dar2020prior,wang2022dimension}.

Quantitative magnetic resonance imaging (qMRI) provides quantitative measures of the physical parameters of tissues, providing additional information regarding its microstructural environment. This is typically accomplished by modeling the acquired MR signal and extracting the parameter of interest. To sufficiently characterize the signal model requires multiple acquisitions, making it both time-consuming and costly, even with well-established acceleration methods. For example, quantifying the spin-lattice relaxation time ($T_1$) using the variable flip angle (vFA) model\cite{FRAM1987201,wang1987optimizing} requires acquisitions at multiple flip angles, leading to impractical scan times for clinical settings.
Recent advances in deep learning have enabled innovative solutions to accelerate qMRI. Methods such as MANTIS\cite{liu2019mantis,liu2020high}, RELAX\cite{liu2021magnetic}, MoDL-QSM\cite{feng2021modl}, and DOPAMINE\cite{jun2021deep} have utilized supervised or self-supervised learning to enable rapid MR parameter mapping using undersampled k-space data.

The desire for more robust and efficient techniques in both MRI and qMRI has led to the development of innovative approaches, among which diffusion models\cite{chung2022come,xie2022measurement,chung2022score,song2022pseudoinverse,gungor2023adaptive} have recently shown to be promising. A notable advancement in this area is the emergence of Denoising Diffusion Probabilistic Models (DDPMs)\cite{ddpm,nichol2021improved,song2020denoising}. These models represent a new class of generative models that have achieved high sample quality without the need of adversarial training. DDPMs have quickly gained interest in MRI reconstruction due to its robustness, especially under distribution shifts. A few studies have explored 
the concept of DDPM-based MRI reconstructions\cite{gungor2023adaptive,chung2022score}. In these methods, DDPMs are trained to generate noisy MRI images, and reconstruction is achieved by iteratively learning to denoise at each diffusion step. This can be implemented through either unconditional or conditional DDPMs. However, to the best of our knowledge, no studies have investigated diffusion models for qMRI.

While diffusion methods have shown promising results, they are not without challenges. Some limitations include the omission of physical constraints during training\cite{ozbey2023unsupervised}, lack of compromise solution and optimal efficiency due to the sampling being initiated from a random noise image \cite{luo2023bayesian}, and reliance on fully-sampled images for training\cite{peng2022towards}. 
In particular, DDPMs for MRI reconstruction generally starts in the image domain, where the unknown data distribution from training images go through repeated Gaussian diffusion processes which can be reversed by learned transition kernels. Applying the diffusion process in the image domain overlooks the underlying MRI physical model (i.e. k-space encoding) which is embedded in the measured k-space data. It can underperform at large distribution shifts, due to changes in scan parameters or difference in anatomy between training and testing. Since the raw MRI measurement is acquired in the frequency domain (i.e. k-space), it can be beneficial to directly apply the diffusion process in the frequency domain rather than image domain. Likewise, since qMRI focuses on the quantification of tissue parameters such as $T_1$ and proton density, it can also be advantageous to define the diffusion model conditioned on the parameter domain for qMRI.

In this paper, we propose a novel and unified method that applies domain-conditioned diffusion models to accelerated static MRI and qMRI reconstruction, which we denote as Static \underline{Di}ffusion \underline{Mo}deling (DiMo) and Quantitative DiMo, respectively. The conditional diffusion process is defined in k-space for Static DiMo and in parameter space for Quantitative DiMo.

Three points that distinguish our method from previous works are: (1) The forward (diffusion) process and reverse (sampling) process are defined on the native data domain rather than image domain. This model is applied to multicoil static MRI and quantitative MRI reconstruction, which reflects domain-specific adaptation. (2) Prior physics knowledge is embedded in the diffusion process as a data consistency component for characterizing MRI k-space encoding and MR signal modeling for MRI and qMRI reconstruction. Gradient descent is also integrated in the diffusion steps to augment feature learning and promote denoising. (3) The proposed method preserves high reconstruction accuracy and efficiency under large undersampling rates for both static MRI and quantitative MRI reconstruction of different anatomies.

The rest of the paper is organized as follows: Section II gives the methods for Static DiMo  and Quantitative DiMo. Section III describes experiment settings. Section IV presents experiment results. Section V discusses the limitation of the method and concludes the paper.

\section{Methods}
\subsection{DDPM}
DDPMs are generative models which are highly effective in learning complex data distributions. The forward diffusion process adds noise to the input data, gradually increasing the noise level until the data is transformed into pure Gaussian noise. This process systematically perturbs the structure of the data distribution. The reverse diffusion process, also known as the denoising process, is then applied to recover the original structure of the data from the perturbed data distribution.

\textbf{Forward Process} DDPM\cite{ddpm} presents the forward diffusion mechanism as a Markov Chain, wherein Gaussian noise is incrementally introduced across several steps to yield a collection of perturbed samples. Consider the uncorrupted data distribution that is characterized by density $q(x_0)$, which undergoes incremental transformations through the addition of Gaussian noise at various stages, resulting in a spectrum of modified samples.  If we draw a data sample $x_0$ from $q(x_0)$, the forward diffusion process modifies this data point through integration of Gaussian perturbations at each time step $t,$ which can be mathematically represented as
\begin{subequations}
\begin{align}
& q(x_{1:T}|x_0) := \prod^T_{t=1} q(x_t|x_{t-1}), \\
& q(x_t | x_{t-1}) := \N(x_t; \sqrt{1-\beta_t} x_{t-1}, \beta_t \II)      
\end{align}
\end{subequations}

Here, $T$ is the total number of diffusion steps, while the noise scaling sequence $\beta_1,...,\beta_T \in [0, 1)$ defines the magnitude of variance introduced at each step. For isotropic Gaussian noise  $\epsilon \sim \N(\textbf{0},\II)$, at each time step, 
\begin{equation}
    x_{t} = \sqrt{1-\beta_t} x_{t-1} + \sqrt{\beta_t }\epsilon.
\end{equation}
With a large number of forward iterations, $x_t$  converges to an isotropic Gaussian sample, so we have $ q(x_T) \approx \N(x_T; \textbf{0}, \II)$.
Let $ \alpha_t :=1-\beta_t$ and $\bar{\alpha}_t := \prod^t_{s=0} \alpha_s$, then given the initial sampled data $x_0$ and sampling step $t$, we can obtain
\begin{equation}\label{diffuser}
    x_{t}( x_{0}, \epsilon) = \sqrt{\bar{\alpha}_t} x_{0} + \sqrt{1- \bar{\alpha}_t }\epsilon.
\end{equation}

\textbf{Reverse Process} The reverse process aims to retrieve $x_0$ by stripping away the noise from $x_T$. Specifically, starting with a noise vector $x_T \sim \N(x_T; \textbf{0}, \II)$, we iteratively sample from the learnable transition kernel $x_{t-1} \sim p_\theta(x_{t-1} | x_t)$ until $t=1$ through a learnable Markov chain in the reverse time direction. 
\begin{subequations}
\begin{align}
& p_\theta(x_{0:T}) := p(x_T)\prod^T_{t=1} p_\theta(x_{t-1} | x_t), \\
& p_\theta(x_{t-1} | x_t) := \N(x_{t-1}; \mu_\theta(x_t,t),\sigma_t^2\II)    \label{reverse}  
\end{align}
\end{subequations}
The mean $\mu_\theta(x_t,t)$ is parametrized by deep neural networks and
$\sigma_t^2 = \frac{1-\bar{\alpha}_{t-1}}{1-\bar{\alpha}_t} \beta_t$. The objective during training is to minimize the discrepancy between the true data distribution and the estimated distribution, which can be achieved by minimizing the  variational bound on log likelihood:
\begin{equation}\label{loglike}
L_{vb} = \EE_{q(x_{0:T})}\Big[ -\log p(x_T) - \sum_{t\ge 1}\log \frac{p_\theta(x_{t-1}|x_t ) }{q(x_t|x_{t-1})} \Big]
\end{equation}
Ho et al.\cite{ddpm} proposed using deep neural networks to estimate added noise $\epsilon$ and perform parameterization on $ \mu_\theta$:
\begin{equation}
    \mu_\theta(x_t, t) = \frac{1}{\sqrt{\alpha_t}} (x_t - \frac{\beta_t}{\sqrt{1-\bar{\alpha}_t}} \epsilon_\theta(x_t, t)),
\end{equation}
Accordingly, the loss function simplifies to 
\begin{equation}
   L_{\text{simple}} = \EE_{x_t, t, \epsilon} \| \epsilon - \epsilon_\theta (x_t , t) \|
\end{equation}

\subsection{Static DiMo: Accelerated MRI reconstruction}
Accelerated MRI considers the following measurement model 
\begin{equation}
    f = Ax + \varepsilon,
\end{equation}
where $x \in \CC^n$ is the MR image of interest consisting of n pixels, $f \in \CC^m$ is its corresponding undersampled k-space measurement, and $\varepsilon \in \CC^n $ is measurement noise. The parameterized forward measurement matrix $ A \in \CC^{m\times n}$ is defined as 
\begin{equation}
    A := \Pcal_\Omega \F \Scal,
\end{equation}
where $\Scal := [\Scal_1, ..., \Scal_j]$ are the coil sensitivity maps for $ j$ different coils, $\F \in \CC^{n \times n}$ is the 2D discrete Fourier transform, and $\Pcal_\Omega \in \NN^{u \times n}  (u \ll n)$ is the binary undersampling mask corresponding to $u$ sampled data points using undersampling pattern $\Omega$. The classical approach to solve for $x$ 
using undersampled k-space $f$ is to solve the following optimization problem:
\begin{equation} \label{ls}
    \min_x  \frac{1}{2} \|  Ax - f \|^2_2.
\end{equation}
The training of Static DiMo starts with the input of fully sampled k-space data, which we denote as $\hat{f}$. We can write $x= \F^{-1} \hat{f}  $, where $ \F^{-1} $ is the inverse discrete Fourier transform. The task of Static DiMo is to minimize   $\hat{f}$, thus we can rewrite equation \eqref{ls} as:
\begin{equation}
    \min_{\hat{f}} \frac{1}{2} \| A\F^{-1} \hat{f} - f \|^2_2
\end{equation}
The minimizer $\hat{f}$ can be solved by applying the gradient decent algorithm, which is straightforward to compute:
\begin{equation}\label{gd}
    \hat{f}^{(k+1)} = \hat{f}^{(k)} - \eta_k \nabla_{\hat{f}^{(k)}}  \frac{1}{2} \| A\F^{-1} \hat{f}^{(k)} - f \|^2_2
 \end{equation}

Typical diffusion models aim to estimate $ q(x | \Pcal_\Omega, f, \Scal)$. Since Static DiMo is defined in the frequency domain and $ x = \F^{-1} \hat{f}$, the problem can be interpreted as generating samples of $q(\hat{f} | \Pcal_\Omega, f, \Scal)$ which is conditioned on undersampling mask $\Pcal_\Omega$, scanned measurement $f$ and coil sensitivity maps $\Scal$.

According to  DDPM\cite{ddpm}, the diffusion process can be represented with a noise schedule $\beta_1,...,\beta_T $ as:
\begin{small}
\begin{subequations} 
\begin{align} 
& q(\hat{f}_{1:T}| \hat{f}_0, \Pcal_\Omega, f, \Scal)  := \prod^T_{t=1} q(\hat{f}_{t}| \hat{f}_{t-1}, \Pcal_\Omega, f, \Scal), \\
& q(\hat{f}_{t}| \hat{f}_{t-1}, \Pcal_\Omega, f, \Scal) := \N(\hat{f}_{t}; \sqrt{ 1- \beta_t} \hat{f}_{t-1}, \beta_t \II ).
\end{align}
\end{subequations}  
\end{small}
Let $ \alpha_t = 1-\beta_t, \bar{\alpha}_t = \sum^t_{s=1} \alpha_s$. We can track the forward process conditioned on the initial: 
\begin{small}
\begin{subequations} 
\begin{align} 
& q(\hat{f}_{t}| \hat{f}_0, \Pcal_\Omega, f, \Scal)  = \N(\hat{f}_{t};\sqrt{\bar{\alpha}_t} \hat{f}_0, (1-\bar{\alpha}_t) \II) \\
& q(\hat{f}_{t-1}| \hat{f}_{t}, \hat{f}_0, \Pcal_\Omega, f, \Scal) = \N(\hat{f}_{t-1} ; \tilde{\mu}_t(\hat{f}_{t}, \hat{f}_0), \tilde{\beta}_t \II ),\label{qt-1} \\ 
& \text{ where }  \tilde{\mu}_t(\hat{f}_{t}, \hat{f}_0) := \frac{ \sqrt{\alpha_t} (1-\bar{\alpha}_{t-1} )}{1-\bar{\alpha}_{t} } \hat{f}_t + \frac{ \sqrt{\bar{\alpha}_{t-1}} \beta_t}{1- \bar{\alpha}_t} \hat{f}_0 \\  \label{posterior}
& \text{ and } \tilde{\beta} := \frac{1- \bar{\alpha}_{t-1} }{1-\bar{\alpha}_t} \beta_t. 
\end{align}
\end{subequations}  
\end{small}
We choose $\alpha_t$ so that $\bar{\alpha}_T \approx 0$, then we obtain 
\begin{equation}
    q(\hat{f}_T | \hat{f}_0, \Pcal_\Omega, f, \Scal ) \approx \N(\hat{f}_T; \textbf{0},  \II),  
\end{equation}
which approaches an isotropic Gaussian distribution. \\
Sample $\hat{f_0} \sim q(\hat{f_0})$, the diffusion model has the following form:  
\begin{small}
\begin{equation}
    p_\theta(\hat{f_0}|  \Pcal_\Omega, f, \Scal) = \int p_\theta (\hat{f}_{0:T} |  \Pcal_\Omega, f, \Scal) \dd_{\hat{f}_{1:T}},
\end{equation}
\end{small}
The sampling of $ p_\theta(\hat{f_0}|  \Pcal_\Omega, f, \Scal) $ is the reverse diffusion process.
\begin{small}
\begin{subequations}
\begin{align} 
& p_\theta(\hat{f}_{0:T}|  \Pcal_\Omega, f, \Scal)  := p_\theta(\hat{f}_T | \Pcal_\Omega, f, \Scal) \prod_{t=1}^{T} p_\theta(\hat{f}_{t-1}| \hat{f}_{t}, \Pcal_\Omega, f, \Scal ), \\
& p_\theta(\hat{f}_{t-1}| \hat{f}_{t}, \Pcal_\Omega, f, \Scal ) := \N(\hat{f}_{t-1}; \mu_\theta( \hat{f}_t, t, \Pcal_\Omega, f, \Scal), \sigma_t^2\II),\label{pt-1}
\end{align}
\end{subequations}
\end{small}
The training of $p_\theta(\hat{f_0}|  \Pcal_\Omega, f, \Scal)$ is to optimize the variational bound of log likelihood in \eqref{loglike}, and this can break down into optimizing random terms of $L$ with stochastic gradient descent. Now we analyse $L_{t-1}$  using  \eqref{qt-1} and \eqref{pt-1}, we can write
\begin{small}
\begin{subequations}
\begin{align}
 L_{t-1} & = D_{KL} (q(\hat{f}_{t-1}|\hat{f}_{t}, \hat{f}_0 , \Pcal_\Omega, f, \Scal) || p_\theta(\hat{f}_{t-1}|\hat{f}_{t},\Pcal_\Omega, f, \Scal) ) \\
 & =  \EE_q[\frac{1}{2\sigma^2} \| \tilde{\mu}_t(\hat{f}_t,\hat{f}_0) - \mu_\theta(\hat{f}_t, t,\Pcal_\Omega, f, \Scal) \|^2_2]  + C,
\end{align}
\end{subequations}
\end{small}
where $C$ is constant that independent from $\theta$. From \eqref{diffuser} we get $ \hat{f}_0 =  \frac{1}{ \sqrt{\alpha_t}} (\hat{f}_t(\hat{f}_0 , \epsilon) - \sqrt{1-\bar{\alpha}_t}\epsilon )$, then apply the forward posterior formula \eqref{posterior}, we have
\begin{small}
\begin{equation}
\begin{aligned}
& L_{t-1} - C  = \EE_q[\frac{1}{2\sigma^2} \|  \frac{1}{ \sqrt{\alpha_t}}(\hat{f}_t(\hat{f}_0 , \epsilon) - \frac{\beta_t}{\sqrt{1-\bar{\alpha}_t}} \epsilon ) -\\
& \mu_\theta(\hat{f}_t(\hat{f}_0 , \epsilon), t,\Pcal_\Omega, f, \Scal) \|^2_2] .
\end{aligned}
\end{equation}
\end{small}
We can derive that $\mu_\theta$ should be parametrized to predict $ \frac{1}{\sqrt{\alpha_t}}( \hat{f}_t - \frac{\beta_t}{\sqrt{1- \bar{\alpha}_t }} \epsilon)$, so we can put
\begin{small}
\begin{equation}\label{mu}
    \mu_\theta(\hat{f}_t, t,\Pcal_\Omega, f, \Scal)  =  \frac{1}{ \sqrt{\alpha_t}} ( \hat{f}_t - \frac{\beta_t}{\sqrt{1- \bar{\alpha}_t} } \epsilon_{\theta}(\hat{f}_t, t, \Pcal_\Omega, f, \Scal)),
\end{equation}
\end{small}
where $\hat{f}_t = \sqrt{\bar{\alpha}_t} \hat{f}_0 + \sqrt{1 - \bar{\alpha}_t} \epsilon, \epsilon \sim \N(\textbf{0}, \II )$.  
Then the simplified loss function is 
\begin{small}
\begin{equation}\label{static_loss}
    \EE_{\hat{f}_0, t, \epsilon} \| \epsilon - \epsilon_\theta( \sqrt{\bar{\alpha}_t} \hat{f}_0 + \sqrt{ 1- \bar{\alpha}_t} \epsilon,  t, \Pcal_\Omega, f, \Scal) \|^2_2, 
\end{equation}
\end{small}
In equation \eqref{static_loss},
we can interpret it as $\hat{f}_t( \sqrt{\bar{\alpha}_t} \hat{f}_0 + \sqrt{ 1- \bar{\alpha}_t} \epsilon,  \Pcal_\Omega, f, \Scal) $ being a function that is dependent on the components of $\epsilon_\theta$. Then we can write our simplified loss function as 
\begin{small}
\begin{subequations}
\begin{align}
L & =  \EE_{\hat{f}_0, t, \epsilon}   \| \epsilon - \epsilon_\theta(\sqrt{\bar{\alpha}_t} \hat{f}_0 + \sqrt{ 1- \bar{\alpha}_t} \epsilon, \Pcal_\Omega, f, \Scal , t) \|^2_2 \\
     & = \EE_{\hat{f}_t, t, \epsilon} \| \epsilon - \epsilon_\theta(\hat{f}_t, t) \|^2_2.
\end{align}
\end{subequations}
\end{small}
Now we can derive the function $\hat{f}_t$ so that it can provide a prior guidance for the diffusion during training and sampling.
The diffusion process starts with $\hat{f}_0 \sim q( \hat{f})$,  we have 
$ \hat{f}_t = \sqrt{\bar{\alpha}_t} \hat{f}_{0} + \sqrt{1- \bar{\alpha}_t }\epsilon$ by \eqref{diffuser}. An intuitive way of designing the function $\hat{f}_t$ is to incorporate a data consistency (DC) layer which perturbs the k-space data $\hat{f}_t$ with a linear combination of partially scanned data $f$.  Then we can further fine-tune the DC layer using gradient descent (GD) as described in \eqref{gd}. 
The proposed algorithm  synergizes  DDPM, MRI data consistency enforcement and gradient descent optimization, which is listed in Alg. \ref{alg:static train}:


 \begin{algorithm}
 \caption{Training Process of Static DiMo}\label{alg:static train}
 \begin{algorithmic}[1]
 \renewcommand{\algorithmicrequire}{\textbf{Input:}}
 \REQUIRE $ t \sim \text{Uniform}(\{ 1,\cdots, T\}), \epsilon \sim \N(\textbf{0}, \II), \hat{f}_0 \sim q(f_0)  $, undersampling mask $\Pcal_\Omega$, partial scanned k-space $f$ and coil sensitivities $\Scal$.
 \\ \textit{Initialisation} : $\eta_0$
\STATE $ \hat{f}_t \leftarrow  \sqrt{\bar{\alpha}_t} \hat{f}_{0} + \sqrt{1- \bar{\alpha}_t }\epsilon.$ 
\STATE $ \hat{f}_t \leftarrow \Pcal_\Omega(\lambda_t f +(1-\lambda_t)\hat{f}_t ) + (\mathbbm{1} - \Pcal_\Omega) \hat{f}_t  $  
\hfill$\triangleright$ DC
\FOR {$k = 1$ to $K$}  
\STATE $ \hat{f}_t \leftarrow \hat{f}_t - \eta_k \nabla_{\hat{f}_t}  \frac{1}{2}\| A \F^{-1} \hat{f}_t - f\|^2_2 $ 
\hfill$\triangleright$ GD
\ENDFOR
\STATE Take gradient descent update step\\
\hspace{5pt} $ \nabla_{\theta} \| \epsilon  - \epsilon_{\theta} (\hat{f}_t ,t ) \|^2_2 $\\
\textbf{Until} converge\\
\renewcommand{\algorithmicensure}{\textbf{Output:}}
\ENSURE  $\hat{f}_t$, $t \in \{ 1,\cdots, T\}$. 
 \end{algorithmic} 
 \end{algorithm}
In step 2, $\lambda_t$ is a scheduling factor to guide the data consistency dynamics, which follows the exponential function to schedule the perturbation dynamics: 
\begin{equation}
    \lambda_t  = \text{exp} (-(t-1) /(T/10)).
\end{equation}
Prior knowledge of scanned signal decreases exponentially as $t$ grows. In doing this, the schedule parameter can balance network training and stabilize parameter learning.
$\mathbbm{1}$ denotes the matrix with values of one and $\eta_k$ is the learnable step size for gradient descent.

Sampling $ \hat{f}_{t-1} \sim p_\theta(\hat{f}_{t-1} |\hat{f}_{t}, \Pcal_\Omega, f, \Scal)$ leverages the Langevin dynamics with $\epsilon_\theta $ as learned gradient of data density  which provide the local information of the data distribution. Here we  compute $ \hat{f}_{t-1} = \frac{1}{ \sqrt{\alpha_t}} ( \hat{f}_t - \frac{\beta_t}{\sqrt{1- \bar{\alpha}_t} } \epsilon_{\theta}(\hat{f}_t, t, \Pcal_\Omega, f, \Scal))+ \sigma_t z $ where $z \sim \N(\textbf{0}, \II)$.
Alg. \ref{alg:static sample} shows the sampling process which is the reverse of the diffusion process.  
 \begin{algorithm}
 \caption{Sampling Process of Static DiMo}\label{alg:static sample}
 \begin{algorithmic}[1]
 \renewcommand{\algorithmicrequire}{\textbf{Input:}}
 \REQUIRE $ \hat{f}_T \sim \N(\textbf{0}, \II)$, undersampling mask $\Pcal_\Omega$, partial scanned k-space $f$ and coil sensitivities $\Scal$.\\
\FOR{$t=T,...,1$}
\STATE $z \sim \N(\textbf{0},\II)$ if  $t>1$, else $z=0$
\STATE $ \hat{f}_{t-1} = \mu_\theta(\hat{f}_t,t ) + \sigma_t z$

\STATE $ \hat{f}_{t-1}  \leftarrow \Pcal_\Omega(\lambda_{t-1}  f +(1-\lambda_{t-1} )\hat{f}_{t-1}  ) + (\mathbbm{1} - \Pcal_\Omega) \hat{f}_{t-1}   $  
\FOR {$k = 0$ to $K$}
\STATE $ \hat{f}_{t-1}  \leftarrow \hat{f}_{t-1} - \eta_k \nabla_{\hat{f}_{t-1} }  \frac{1}{2}\| A \F^{-1} \hat{f}_{t-1} - f\|^2_2 $
\ENDFOR
\ENDFOR
\renewcommand{\algorithmicensure}{\textbf{Output:}}
\ENSURE  $\hat{f}_0$ 
 \end{algorithmic} 
 \end{algorithm}

\subsection{Quantitative DiMo: Accelerated qMRI reconstruction.}
To extend from static MRI reconstruction to qMRI reconstruction, we first define the MR parameter maps as $ \Delta =\{\delta_i \}_{i=1}^N$ where $\delta_i$ symbolizes each MR parameter and $N $ is the total number of MR parameters to be estimated. Given the MR signal model $\Mbf$, which is a function of $\Delta$,  the MR parameters $\Delta $ can be estimated by minimizing the following:
\begin{equation} \label{ls_qMRI}
    \min_{\Delta}  \frac{1}{2} \|  A \Mbf(\Delta)  - f \|^2_2.
\end{equation}
Quantitative DiMo is defined in the parameter domain, therefore  we focus on estimating $q( \hat{\Delta} | \Pcal_\Omega, f, \Scal, \Mbf )$ with an additional condition, the MR signal model $\Mbf$. Since the diffusion processe and reverse process are highly similar to Static DiMo, we omit the derivation of Quantitative DiMo. 

In contrast to the algorithm design of Static DiMo,
a major difference is the function $ \hat{\Delta}_t( \bar{\alpha}_t \hat{f}_0 + \sqrt{1-\bar{\alpha}_t} \epsilon, \Pcal_\Omega, f, \Scal, \Mbf )$. Following the same framework as Static DiMo, Quantitative DiMo starts with equation \eqref{diffuser}, which is then input to the quantitative DC (QDC) layer, followed by fine-tuning with gradient descent. The Alg. \ref{alg:qMRI train} displays the training process of Quantitative DiMo and Alg. \ref{alg:qMRI sample} shows the sampling process:

 \begin{algorithm}
 \caption{Training Process of Quantitative DiMo} \label{alg:qMRI train}
 \begin{algorithmic}[1]
 \renewcommand{\algorithmicrequire}{\textbf{Input:}}
 \REQUIRE $ t \sim \text{Uniform}(\{ 1,\cdots, T\}), \epsilon \sim \N(\textbf{0}, \II), \hat{\Delta}_{0} \sim q(\Delta_{0})$, and acquired data $\Pcal_\Omega, f, \Scal, \Mbf$. \hspace{19.5mm}\tikzmark{right}
 \\ \textit{Initialisation} : $\tau_0$
\STATE $ \hat{\Delta}_{t} \leftarrow  \sqrt{\bar{\alpha}_t} \hat{\Delta}_{0} + \sqrt{1- \bar{\alpha}_t }\epsilon$
\STATE  $ \hat{f}_{t} \leftarrow  \F \Scal \Mbf(\hat{\Delta}_{t})$ \tikzmark{top}
\STATE $ \hat{f}_t \leftarrow \Pcal_\Omega(\lambda_t f +(1-\lambda_t)\hat{f}_t ) + (\mathbbm{1} - \Pcal_\Omega) \hat{f}_t $
\STATE $ \hat{\Delta}_{t} \leftarrow   \Mbf^{-1} \Scal^{-1}( \F^{-1} \hat{f}_t  )$ \tikzmark{bottom}
\FOR {$k = 0$ to $K$}
\STATE $ \hat{\Delta}_{t} \leftarrow  \hat{\Delta}_{t} - \tau_k  \nabla_{\hat{\Delta}_{t}}  \frac{1}{2}\| A \Mbf (\hat{\Delta}_{t}) - f\|^2_2 $
\hfill$\triangleright$ GD
\ENDFOR
\STATE Take gradient descent update step\\
\hspace{5pt} $ \nabla_{\theta} \| \epsilon  - \epsilon_{\theta} (\hat{\Delta}_t ,t ) \|^2_2 $\\
\textbf{Until} converge\\
\renewcommand{\algorithmicensure}{\textbf{Output:}}
\ENSURE   $\hat{\Delta}_t$, $t \in \{ 1,\cdots, T\}$.
 \AddNote{top}{bottom}{right}{\hfill$\triangleright$ QDC}
 \end{algorithmic} 
 \end{algorithm}
 \begin{algorithm}
\caption{Sampling Process of Quantitative DiMo}\label{alg:qMRI sample}
\begin{algorithmic}[1]
\renewcommand{\algorithmicrequire}{\textbf{Input:}}
\REQUIRE $ \hat{\Delta}_T \sim \N(\textbf{0},\II)$, and acquired data $\Pcal_\Omega, f, \Scal, \Mbf$.
\FOR{$t=T,...,1$}
\STATE $z \sim \N(\textbf{0}, \II)$ if  $t>1$, else $z=0$
\STATE $ \hat{\Delta}_{t-1} = \mu_\theta(\hat{\Delta}_t,t ) + \sigma_t z$
\STATE $ \hat{f}_{t-1} \leftarrow  \F \Scal \Mbf(\hat{\Delta}_{t-1})$ 
\STATE $ \hat{f}_{t-1}  \leftarrow \Pcal_\Omega(\lambda_{t-1}  f +(1-\lambda_{t-1} )\hat{f}_{t-1}  ) + (\mathbbm{1} - \Pcal_\Omega) \hat{f}_{t-1}   $  
\STATE $ \hat{\Delta}_{t-1} \leftarrow   \Mbf^{-1} \Scal^{-1}( \F^{-1} \hat{f}_{t-1}  )$ \\
\FOR {$k = 0$ to $K$}
\STATE $ \hat{\Delta}_{t-1} \leftarrow  \hat{\Delta}_{t-1} - \tau_{k}  \nabla_{\hat{\Delta}_{t-1}}  \frac{1}{2}\| A \Mbf (\hat{\Delta}_{t-1}) - f\|^2_2 $\\
\ENDFOR
\ENDFOR
\renewcommand{\algorithmicensure}{\textbf{Output:}}
\ENSURE  $\hat{\Delta}_0$ 
 \end{algorithmic} 
 \end{algorithm}
QDC layer is presented in steps 2-4 in the above algorithm. The addition of the QDC step is to perturb quantitative parameters and guide the parameter space denoising and generation process. Steps 5-7 further optimize the perturbed $\hat{\Delta}$ through iterating $K$ steps of gradient descent. 
Alg. \ref{alg:qMRI sample} displays the sampling process which is the reverse of the diffusion process.

\begin{figure*}
\includegraphics[width=1\linewidth]{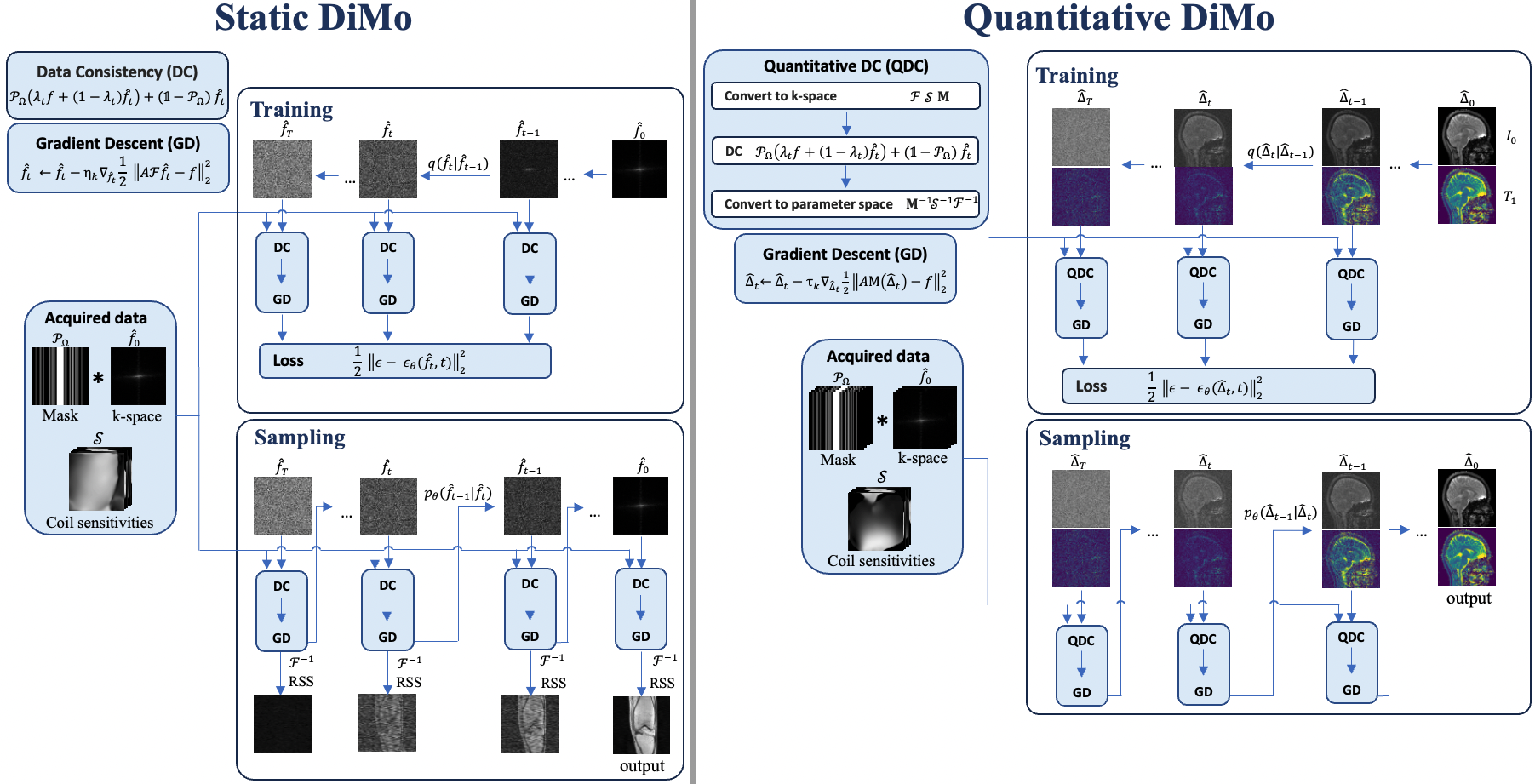}
\caption{Diffusion model framework for Static DiMo and Quantitative DiMo (e.g., $T_1$ mapping).}\label{fig:framework}
\end{figure*}

\begin{figure}
\includegraphics[width=1\linewidth]{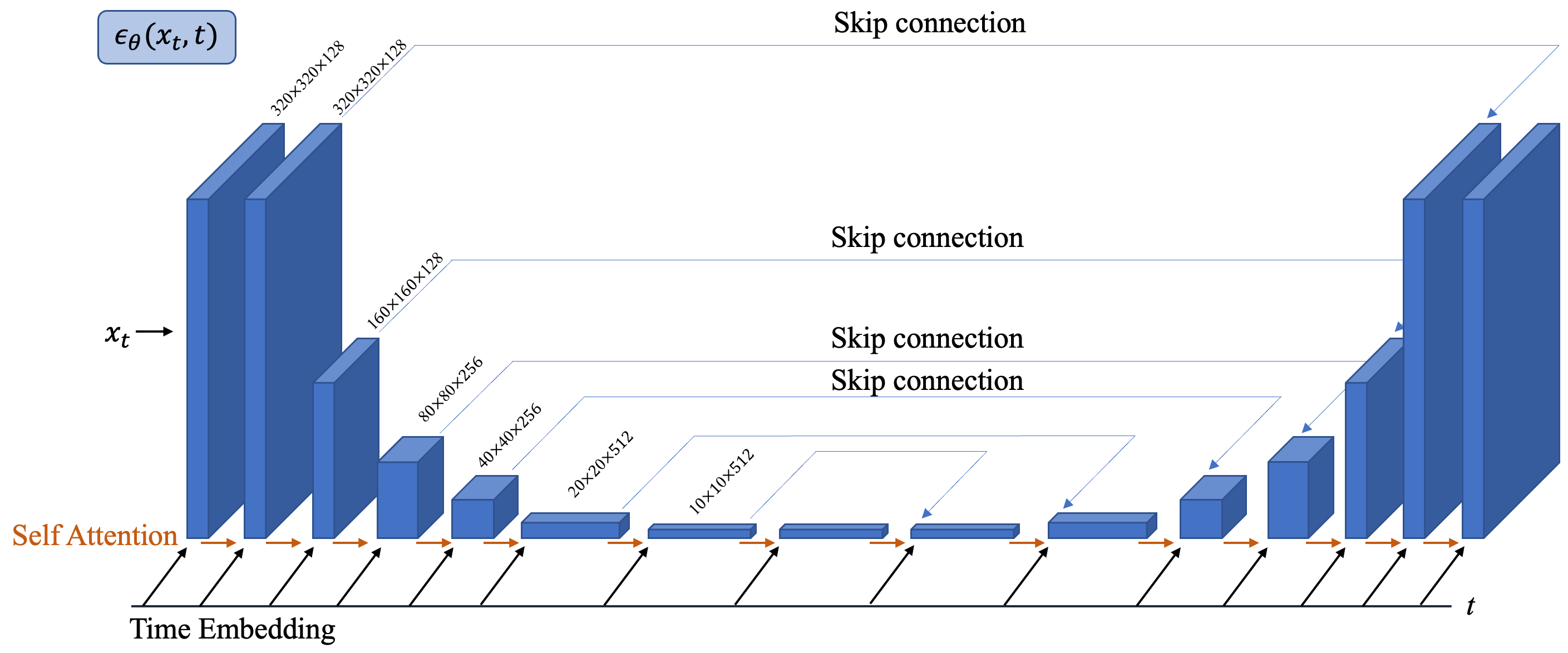}
\caption{Network structure for $ \epsilon_{\theta} (x_t, t) $.}\label{fig:Unet}
\end{figure}

To demonstrate the feasibility of Quantitative DiMo for reconstructing accelerated qMRI, we used $T_1$ mapping using the vFA model \cite{wang1987optimizing} as an example. For MR images are acquired at multiple flip angles  $ \phi_i$, where $i = 1,...,M$,  the MR signal model $\Mbf $ reads as:
\begin{equation}\label{vfa}
    \Mbf_{i}( T_1, I_0) =  I_0 \cdot \frac{(1-e^{-TR/ T_1}) \sin{\phi_i}}{ 1 -e^{-TR/ T_1} \cos{\phi_i}}, 
\end{equation}
where $T_1 \in \RR^n $  denotes the spin-lattice relaxation time map and $I_0 \in \CC^n $ is proton density map. In this model, the MR parameters estimated are encapsulated in the set $ \Delta = \{ T_1,I_0 \}$. Other imaging parameters such as flip angles and repetition time TR are known. 
Fig. \ref{fig:framework} shows the framework of Static DiMo and Quantitative DiMo and Fig. \ref{fig:Unet} illustrates the U-Net architecture for $ \epsilon_\theta(x_t, t)$.

\section{Experiments}

\subsection{Experiments for Static DiMo}
We acquired knee data from 25 subjects, using 20 for training and 5 for testing. These fully-sampled data were obtained from a 3T GE Premier scanner equipped with an 18-element knee coil array. The data was acquired using two-dimensional fast spin-echo (FSE) sequences along the coronal plane, with both proton density-weighting (PD-weighting) and $T_2$-weighting. All images were resized resulting in in-plane matrix size of 320 $\times$ 320 with 3mm  slice thickness. 
Number of slices varied from 30 to 38 in each dataset, amounting to 875 total slices for  both training and testing. The experiments were conducted retrospectively using three different acceleration factors: AFs = 4$\times$, 5$\times$ and 6$\times$ using 1D variable density Cartesian undersampling masks\cite{lustig2007sparse} where the 20
central k-space lines were fully sampled.
We evaluated the performance of the proposed method against several state-of-the-art methods which include Variational Network (VN)\cite{hammernik2018learning}, ISTA-NET\cite{zhang2018ista}, and pMRI-Net\cite{bian2022optimal} over two contrasts: PD- and $T_2$-weighting. Results were compared in terms of peak signal to noise ratio (PSNR), structure similarity (SSIM) and normalized mean squared error (NMSE). The equations for assessing PSNR, SSIM, and NMSE between reconstruction $\vrm$ and the ground truth $\vrm^*$ are:
\begin{equation}
    PSNR(\vrm, \vrm^*) = 20\log_{10} (\max(|\vrm^*|)/\tfrac{1}{N} \parallel \vrm - \vrm^* \parallel^2),
\end{equation}
\begin{equation}
    SSIM(\vrm, \vrm^*) = \frac{(2 \mu_\vrm  \mu_{\vrm^*} +c_1)(2 \sigma_{\vrm \vrm^*} + c_2)}{(\mu_\vrm^2 +\mu_{\vrm^*}^2 + c_1)(\sigma_\vrm^2 + \sigma_{\vrm^*}^2+ c_2)},
\end{equation}
\begin{equation}
    NMSE(\vrm, \vrm^*) = \parallel \vrm^* - \vrm \parallel^2_2 / \parallel \vrm^* \parallel^2_2, 
\end{equation}
where$\mu_\vrm , \mu_{\vrm^*}$  represent the local mean of pixel intensities with standard deviations  $ \sigma_\vrm, \sigma_{\vrm^*}$  for images $\vrm, \vrm^*$, respectively. $\sigma_{\vrm \vrm^*}$   denotes the covariance between $\vrm$ and $\vrm^*$, and $C_1 = (K_1 L)^2, C_2 = (K_2L)^2 $ are two constant variables introduced to stabilize the division.
We chose the total number of sampling steps as $T=1000$ using a linear schedule from $ \beta_1=10^{-5}$ to $\beta_T=10^{-2}$ with training batch size of 4. The total number of epochs used was 7000. The initial parameter for Alg.\ref{alg:static train} was chosen to be $\eta_0=10^{-4}$.

\subsection{Experiments for Quantitative DiMo}
Quantitative DiMo experiments were carried out using in-vivo brain data of healthy volunteers, obtained from a Siemens 3T Prisma scanner equipped with a 20-channel head coil. The five subject fully sampled vFA brain data was acquired along the sagittal plane using a spoiled gradient echo sequence with imaging parameters  TE/TR = $12/40$ ms, FA = $5^\circ, 10^\circ, 20^\circ, 40^\circ$, in-plane matrix size = $176 \times 176$ with 3mm slice thickness and a total number of 48 slices per subject. Leave-One-Out cross-validation was used for all five subjects to train and test our method. 

Two undersampling schemes were retrospectively used: (1) 1D variable density Cartesian 
undersampling\cite{lustig2007sparse} with acceleration factor AF= $4 \times$, where the 16 central k-space lines were fully sampled and (2) 2D Poisson disk undersampling at AF= $4 \times$ with the central 51x51 k-space portion fully sampled. The undersampling patterns were varied for each flip angle, like in previous studies\cite{liu2019mantis,liu2021magnetic}.
We compared Quantitative DiMo with two advanced non-deep learning qMRI reconstruction techniques Locally Low Rank (LLR)\cite{zhang2015accelerating}, 
Model-TGV\cite{maier2019rapid}, 
and a self-supervised deep learning method  RELAX\cite{liu2021magnetic}. 
For LLR and Model-TGV, settings recommended in their original research papers were used along with available code. RELAX and Quantitative DiMo were trained through cross-validation. For the diffusion hyperparameters, we maintained the total number of sampling steps at $T = 1000$ using linear scheduling from  $\beta_1=10^{-6}$ to $\beta_T = 0.05$. The training batch size used was 8 with 5500 total epochs. The initial parameter for gradient descent in Alg. \ref{alg:qMRI train} was set to $ \tau_0 = 10^{-5}.$

All in-vivo studies were carried out under a protocol approved by our institution’s institutional review board.
Another experiment condition was as follows:
The coil sensitivity maps were estimated from ESPIRiT\cite{uecker2014espirit}. Separate  $B_1^+$ maps were acquired\cite{sacolick2010b1} to correct for $B_1^+$ bias in estimating $T_1$. All the programming in this study was implemented using Python language and PyTorch package, and experiments were conducted on one NVIDIA A100 80GB GPU and an Intel Xeon 6338 CPU at Centos Linux system.

\section{Results}
\subsection{Results of Static DiMo}
\begin{figure}
\includegraphics[width=1\linewidth]{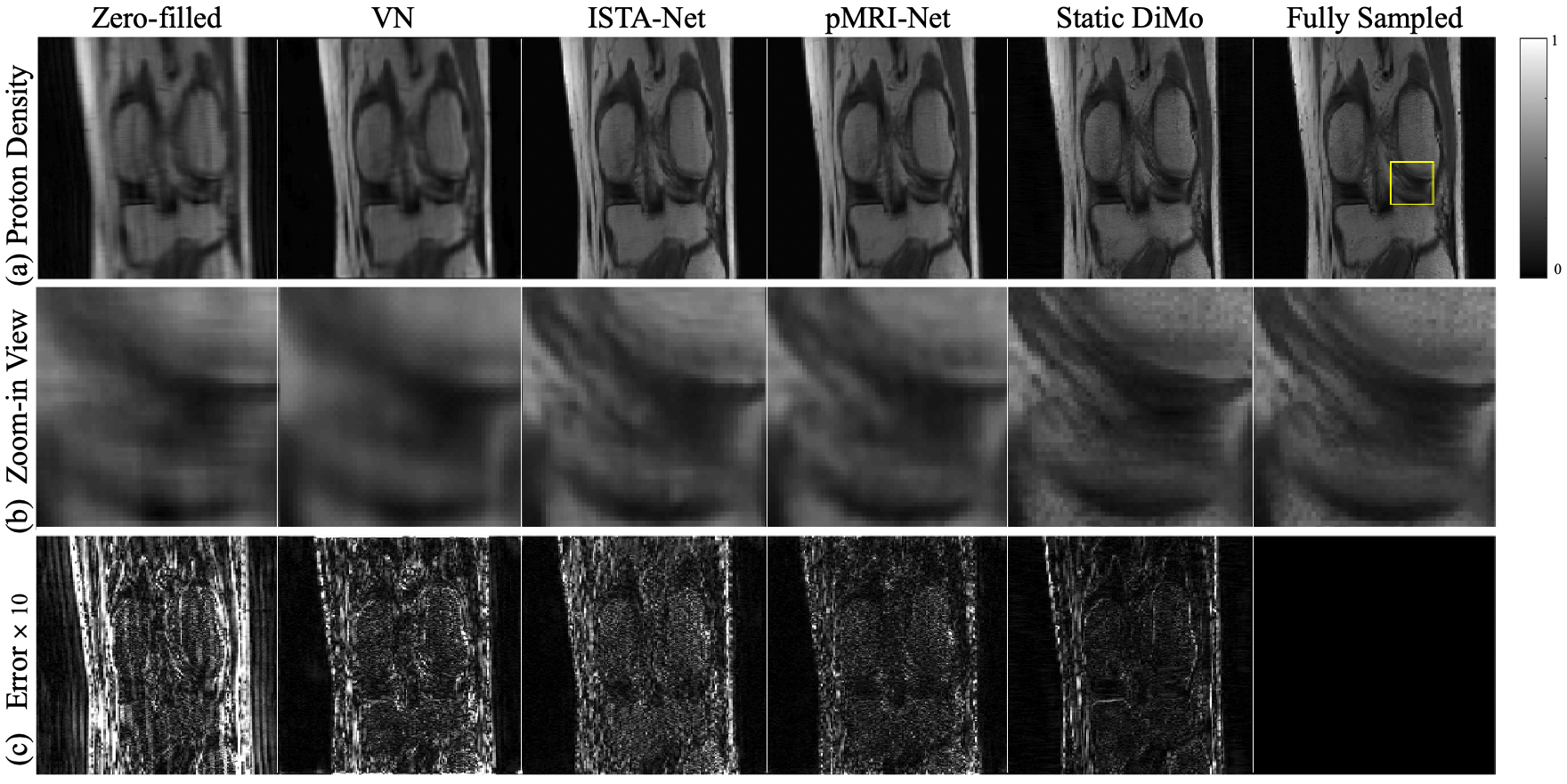}
\caption{Comparison of coronal PD-weighting contrast at AF = 6$\times$.}\label{fig:pd}
\end{figure}

\begin{figure}
\includegraphics[width=1\linewidth]{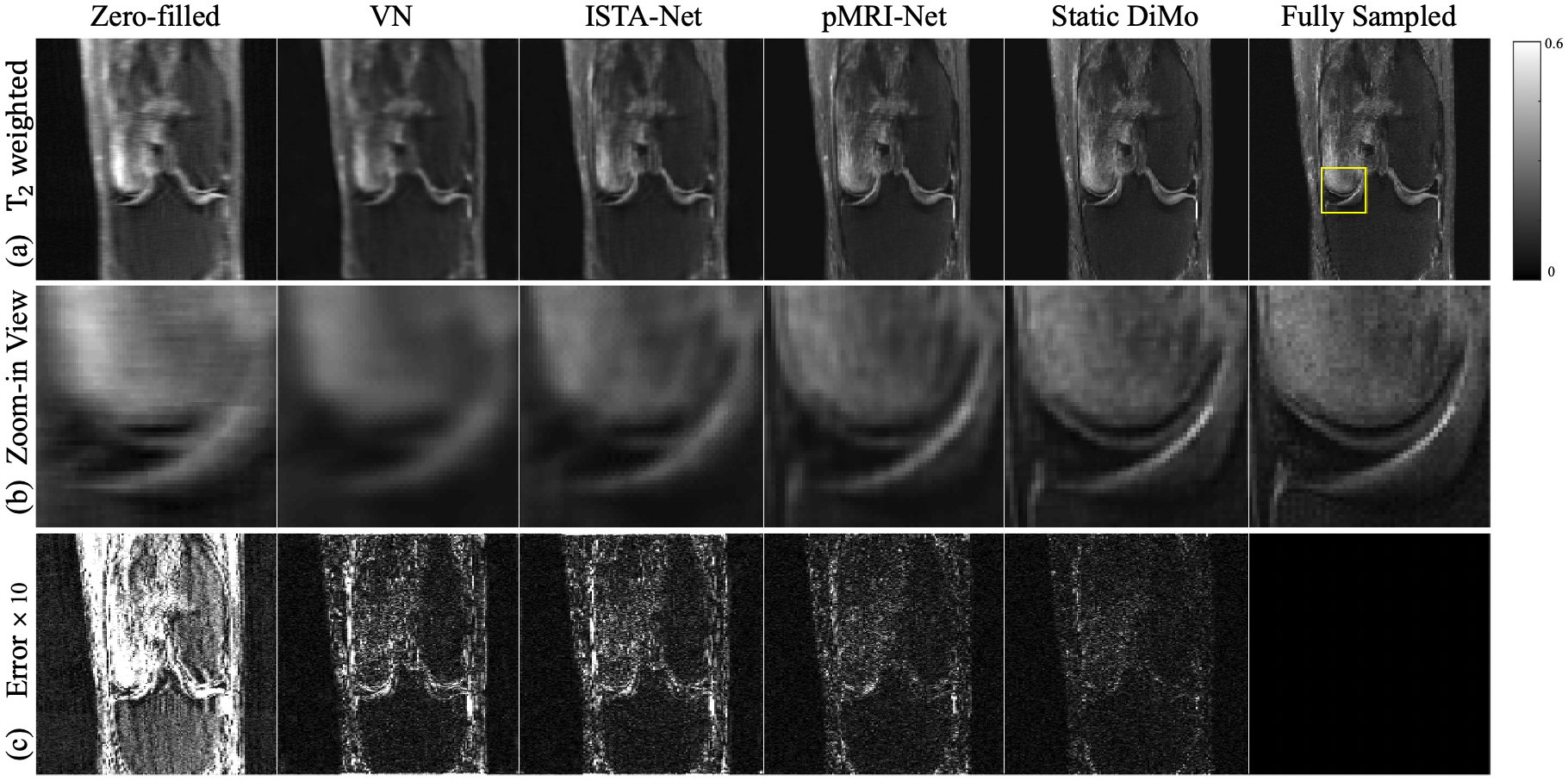}
\caption{Comparison of coronal T2-weighting contrast at AF =  6$\times$.}\label{fig:t2}
\end{figure}

An example of reconstruction results obtained from different methods at AF = $6 \times$ are shown for PD- (Fig. \ref{fig:pd}) and T2-weighted (Fig. \ref{fig:t2}) images along the coronal plane are displayed. Zero-filled results show significant blurring and artifacts due to undersampling. Although VN, ISTA-Net, and pMRI-Net are able to remove these artifacts, compared with the fully sampled reference images, blurring characteristic persists with VN showing the most, while ISTA-Net and pMRI-Net are comparable. Static DiMo's reconstruction not only removes the undersampling artifacts, but also significantly suppresses blurring showing clarity and sharpness, thus outperforming both ISTA-Net and pMRI-Net. Furthermore, Static DiMo shows superior denoising capability, rooted in its denoising diffusion modeling. This not only proficiently removes noise and artifacts, but also retains the integrity of high-frequency image detail. This is further illustrated in the zoomed-in images of Fig. \ref{fig:pd}(b) and Fig. \ref{fig:t2}(b). Static DiMo clearly distinguishes tissue boundaries across various tissue types, including the cartilage, meniscus, bone and muscle. In addition, the fidelity of tissue texture and sharpness are preserved. The pixel-wise error maps shown in Fig. \ref{fig:pd}(c) and Fig. \ref{fig:t2}(c) also demonstrate that Static DiMo produces the least reconstruction errors with respect to the fully sampled reference. For both contrasts, Static DiMo consistently outperforms the other methods. Overall, in agreement with the qualitative observation in Fig. \ref{fig:pd} and Fig. \ref{fig:t2},  reconstruction  results are summarized in Table \ref{static_table} for all testing subjects using quantitative metrics, showing the superior performance of Static DiMo over other methods in reconstruction fidelity, structure and texture preservation and noise suppression.
\begin{table*}[t]
\centering
\caption{Comparison of different methods using quantitative metrics.} \label{static_table}
 \resizebox{\linewidth}{!}{
\begin{tabular}{c|ccc|ccc|ccc}
\hline 
\multicolumn{10}{c}{\textbf{PD-weighting$^1$}} \\
\hline 
AF   & & 4x &  &  & 5x &  &  & 6x &  \\
\hline
Methods &  PSNR & SSIM & NMSE  & PSNR & SSIM & NMSE & PSNR & SSIM & NMSE   \\
  VN \cite{hammernik2018learning}  &   30.5932 $\pm$ 0.9312 & 0.9016 $\pm$ 0.0390  & 0.1374 $\pm $ 0.0203  &  29.3312 $\pm$  1.1914  & 0.8868 $\pm$ 0.0482 & 0.1467 $\pm$ 0.0269   &   27.5036 $\pm$  1.2453 & 0.8503 $\pm$ 0.0594   & 0.1788 $\pm$ 0.0301 \\
  
 ISTA-Net\cite{zhang2018ista} & 32.3933 $\pm$ 1.1124 & 0.9640 $\pm$ 0.0260  & 0.1030 $\pm$ 0.0121  & 31.1878 $\pm$ 1.2343  & 0.9555 $\pm$ 0.0369 & 0.1150 $\pm$  0.0152  &  29.1795 $\pm$ 1.2951 & 0.9603 $\pm$   0.0415 & 0.1281 $\pm$0.0197   \\
 
  pMRI-Net\cite{bian2022optimal} & 33.5534 $\pm$ 1.2351 & 0.9695 $\pm$  0.0144 & 0.0937 $\pm$ 0.0058  & 32.7408 $\pm$  1.4180 & 0.9667 $\pm$ 0.0198 & 0.1003 $\pm$  0.0086  &  30.8576 $\pm$ 1.5202  &  0.9560 $\pm$ 0.0235  & 0.1178  $\pm$   0.0128 \\
  
 Static DiMo & 35.3733 $\pm$ 1.2142  & 0.9704 $\pm$ 0.0137   & 0.0738 $\pm$ 0.0044  &  34.0121 $\pm$ 1.3185  & 0.9691 $\pm$  0.0169 &  0.0839 $\pm$  0.0081 & 32.7645 $\pm$ 1.4719 & 0.9597 $\pm$ 0.0119 & 0.1084 $\pm$ 0.0109 \\
 \hline
\end{tabular} }
 \resizebox{\linewidth}{!}{
\begin{tabular}{c|ccc|ccc|ccc}
\hline   
\multicolumn{10}{c}{ \textbf{$\boldsymbol T_2$-weighting$^1$}} \\
\hline 
Methods &  PSNR & SSIM & NMSE  & PSNR & SSIM & NMSE & PSNR & SSIM & NMSE   \\
VN \cite{hammernik2018learning} &  32.0078 $\pm$ 0.5800  & 0.9129 $\pm$ 0.0119 & 0.1464 $\pm$ 0.0048  &   30.8641 $\pm$0.6303 & 0.8975 $\pm$ 0.0128  &  0.1743 $\pm$ 0.0082  &    29.2356 $\pm$  0.7866 & 0.8708 $\pm$ 0.0245 & 0.1934 $\pm$ 0.0102 \\

ISTA-Net\cite{zhang2018ista} & 34.1552 $\pm$ 0.7873  & 0.9628 $\pm$ 0.0056 & 0.1142 $\pm$ 0.0042 &  32.7243 $\pm$  1.0213 & 0.9472 $\pm$ 0.0093  & 0.1382 $\pm$ 0.0065 &  30.0678 $\pm$ 1.1038  & 0.9155 $\pm$ 0.0109 & 0.1864 $\pm$ 0.0080 \\

pMRI-Net\cite{bian2022optimal}  & 36.3567 $\pm$ 1.1123  & 0.9767 $\pm$ 0.0045  & 0.0867 $\pm$ 0.0032 &  35.5938 $\pm$ 1.1256  & 0.9728 $\pm$ 0.0078  & 0.0942 $\pm$ 0.0045&  34.1399 $\pm$  1.1321 & 0.9646 $\pm$ 0.0087 & 0.1108 $\pm$ 0.0059 \\

Static DiMo & 37.6438 $\pm$ 1.0343 & 0.9817 $\pm$ 0.0039 & 0.0776 $\pm$ 0.0025 &  36.2274 $\pm$ 1.1227 & 0.9812 $\pm$  0.0071 & 0.0793 $\pm$ 0.0049 &  35.1207 $\pm$  1.1348 & 0.9708 $\pm$  0.0079 & 0.0983 $\pm$ 0.0056\\
\hline
\end{tabular}
}
\\[1ex] 
\makebox[\linewidth][c]{%
    \begin{tabular}{lcl}
    \scriptsize $^1$ Data are presented as mean $\pm$ std. 
    \end{tabular}}%
\end{table*}


\subsection{Results of Quantitative DiMo}
\begin{figure*}
\includegraphics[width=1\linewidth]{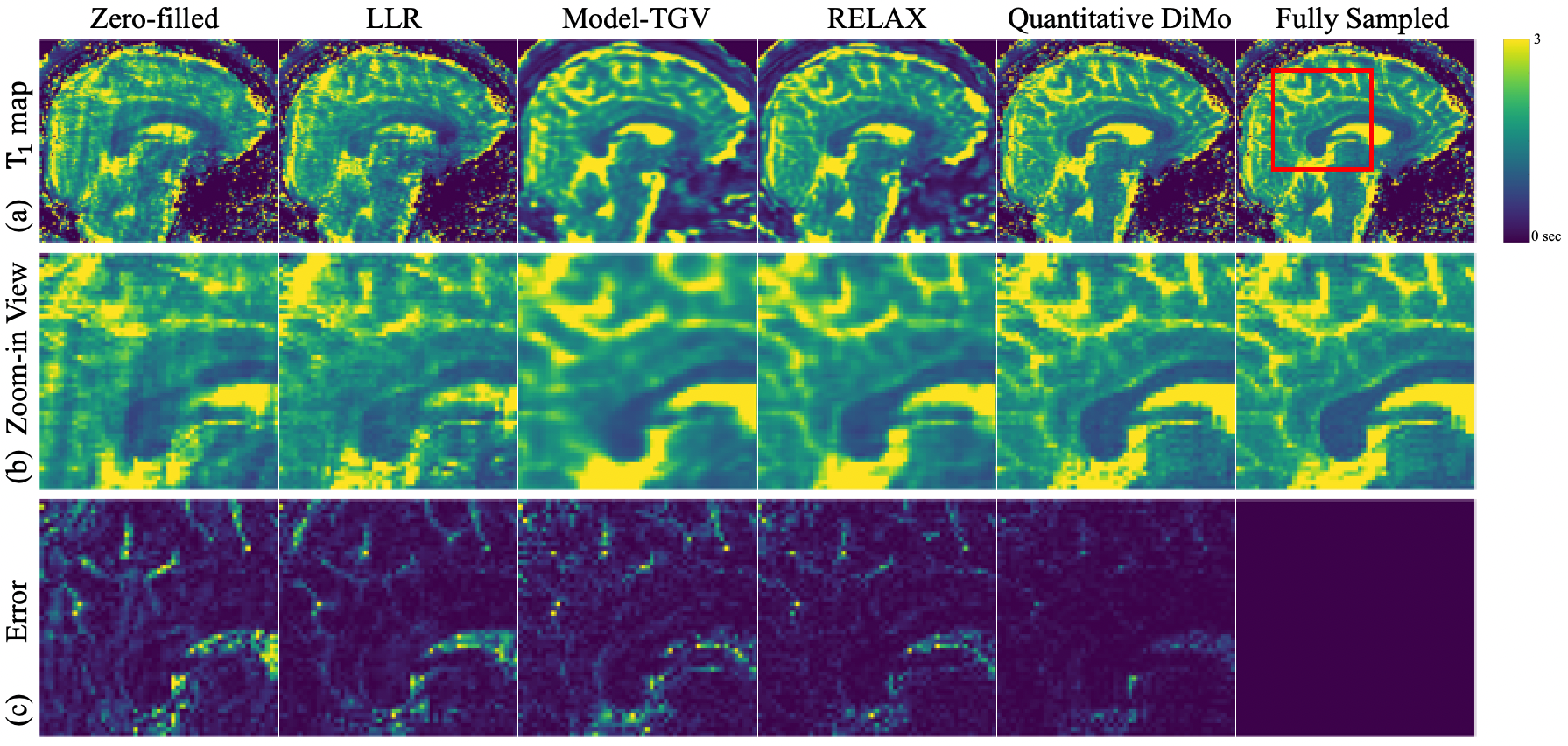}\\
\includegraphics[width=0.986\linewidth]{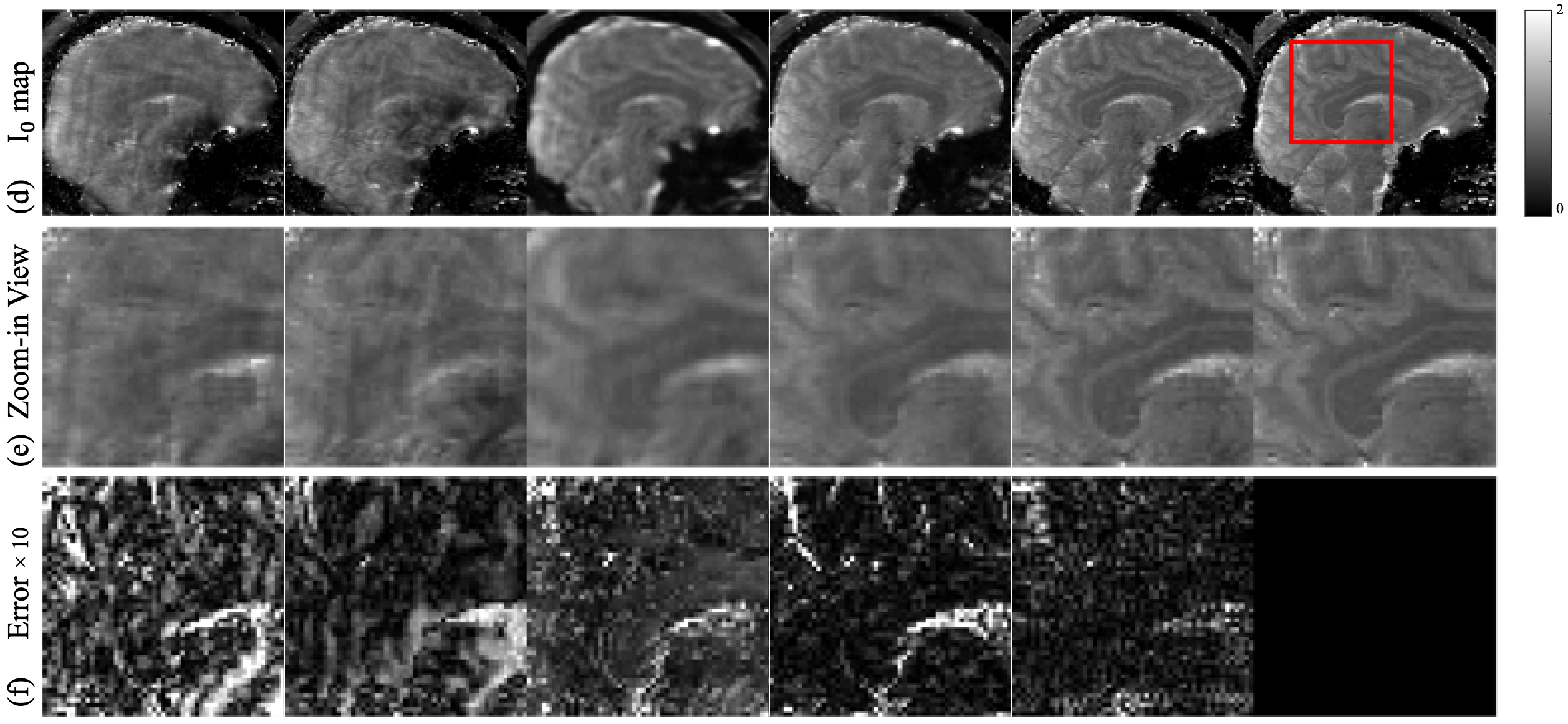}
\caption{Exemplified comparison of $T_1$  and $I_0$ mapping among different methods using 4$\times$ 1D Cartesian variable density undersampling
mask.}\label{fig:cartesian}
\end{figure*}

$T_1$ and $I_0$ maps estimated from 4$\times$ 1D variable density Cartesian undersampling using various methods are shown in Fig. \ref{fig:cartesian} The zero-filled $T_1$ map derived from pixel-wise fitting the undersampled images (Fig. \ref{fig:cartesian}(a)) exhibits noise and ripple artifacts due to aliasing. While LLR manages to mitigate these artifacts to an extent, the noisy signature persists. In its attempt to neutralize the noise, Model-TGV yields a more refined tissue appearance with enhanced $T_1$ contrast. However, the resulting maps are overly smooth, appearing blurry. Conversely, RELAX effectively suppresses both noise and artifacts, delivering sharper maps, albeit with a slightly blocky texture. This blockiness is speculated to arise from challenges in achieving convergence in the end-to-end network using limited data. Meanwhile, Quantitative DiMo generates clear and sharp $T_1$  maps. Its proficiency in removing noise translates to its superior map quality, both in terms of appearance and contrast. This is further witnessed in the zoom-ins (Fig. \ref{fig:cartesian} (b)) which show that Quantitative DiMo clearly distinguishes the boundary between white matter and grey matter, appearing nearly identical to the fully sampled reference map. This is quantitatively confirmed in the error maps (Fig. \ref{fig:cartesian}(c)) where the zero-filled incurs the largest error, followed by LLR, Model-TGV, and RELAX.

Mean $T_1$ values obtained from representative white matter and grey matter regions are presented in Table \ref{Quant_table} for all subjects. Overall, LLR shows least agreement with respect to the fully sampled reference method, followed by Model-GTV. On the other hand, both RELAX and Quantitative DiMo show good agreement. However, as shown by the Wilcoxon signed rank test results, Quantitative DiMo shows better correlation with the reference than RELAX.
\begin{table*}[]
    \centering
    \caption{ROI analysis of representative brain regions for 1D Cartesian undersampling at AF = 4$\times$.} \label{Quant_table}
    \begin{tabular}{l cccccc}
    \hline \vspace{1mm}
     & & & $T_1[s]^1$ & &\\
       Region-of-interest   & LLR & Model-TGV & RELAX & Quantitative DiMo & Fully Sampled \\
       \hline
     \textbf{White matter region}   & & & &  &\\
      Corpus Callosum    & $0.957 \pm 0.217$ & $ 0.902 \pm 0.186$ & $0.895 \pm 0.173$ & $ 0.894 \pm 0.167^{\S}$ & $ 0.895 \pm 0.164$ \\
      Frontal white matter  & $ 0.932 \pm 0.226$ & $ 0.918 \pm 0.213$ & $0.906 \pm 0.207^{\S} $ & $0.905 \pm 0.203^{\S}$ & $0.905 \pm 0.202$ \\
      \textbf{Grey matter region} & & & & &\\
      Putamen & $1.201 \pm 0.209 $ & $ 1.287 \pm 0.175$ & $ 1.278 \pm 0.174$ & $ 1.308 \pm 0.170^{\S}$ & $ 1.311 \pm 0.163$ \\
      Thalamus & $1.252 \pm 0.201$ & $ 1.224 \pm 0.162$ & $ 1.216 \pm 0.134$ & $ 1.223 \pm 0.118$ & $ 1.225 \pm 0.109$\\
      \hline
    \end{tabular}
    \\[1ex] 
    \makebox[\linewidth][c]{%
        \begin{tabular}{lccl} \footnotesize
            $^1$ Data are presented as mean $\pm$ std. &
            $\S$ $P> 0.05$ vs. fully sampled $T_1$ using Wilcoxon signed rank test.
        \end{tabular}}%
    \label{tab:Quant_table}
\end{table*}
\begin{figure*}
\includegraphics[width=1\linewidth]{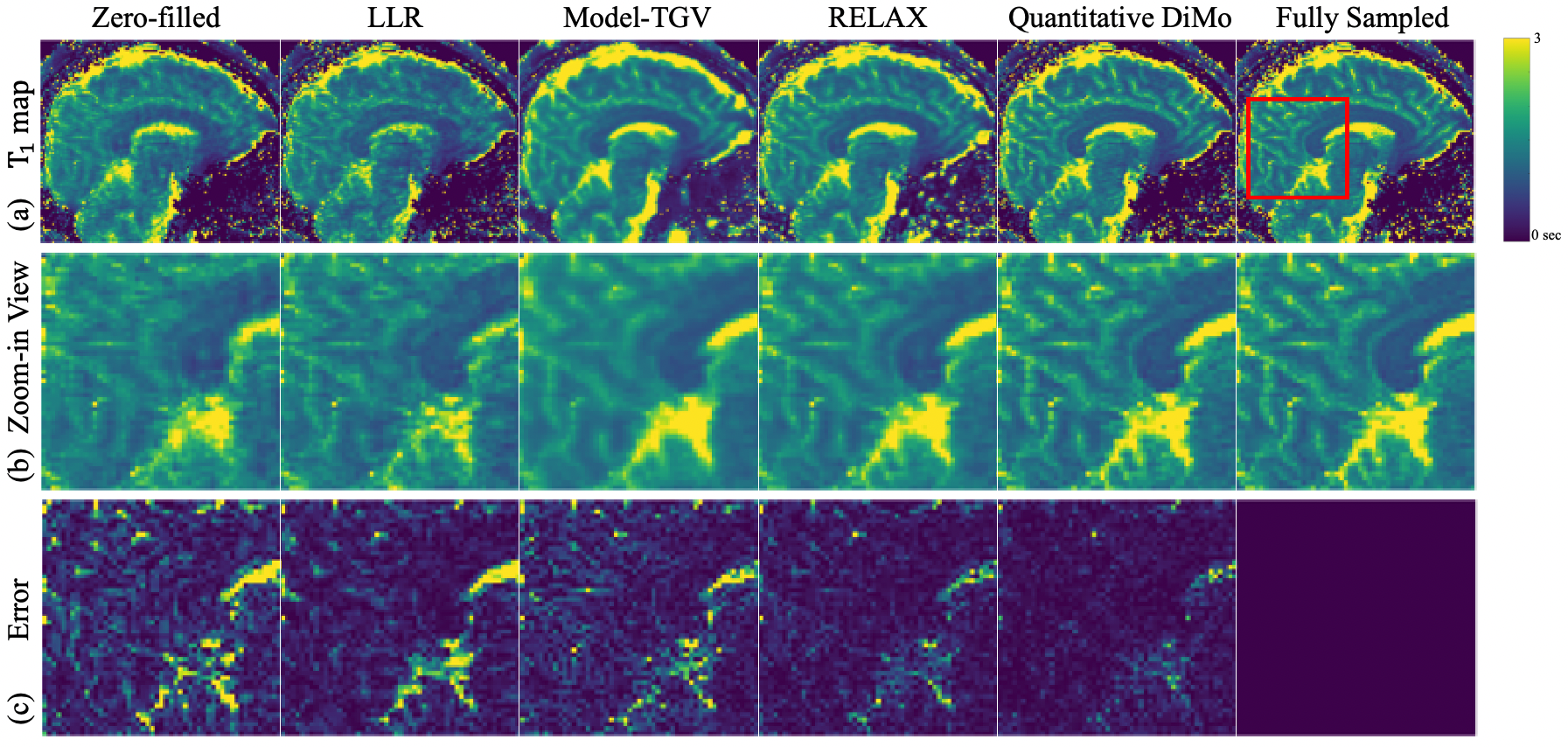}\\
\includegraphics[width=0.986\linewidth]{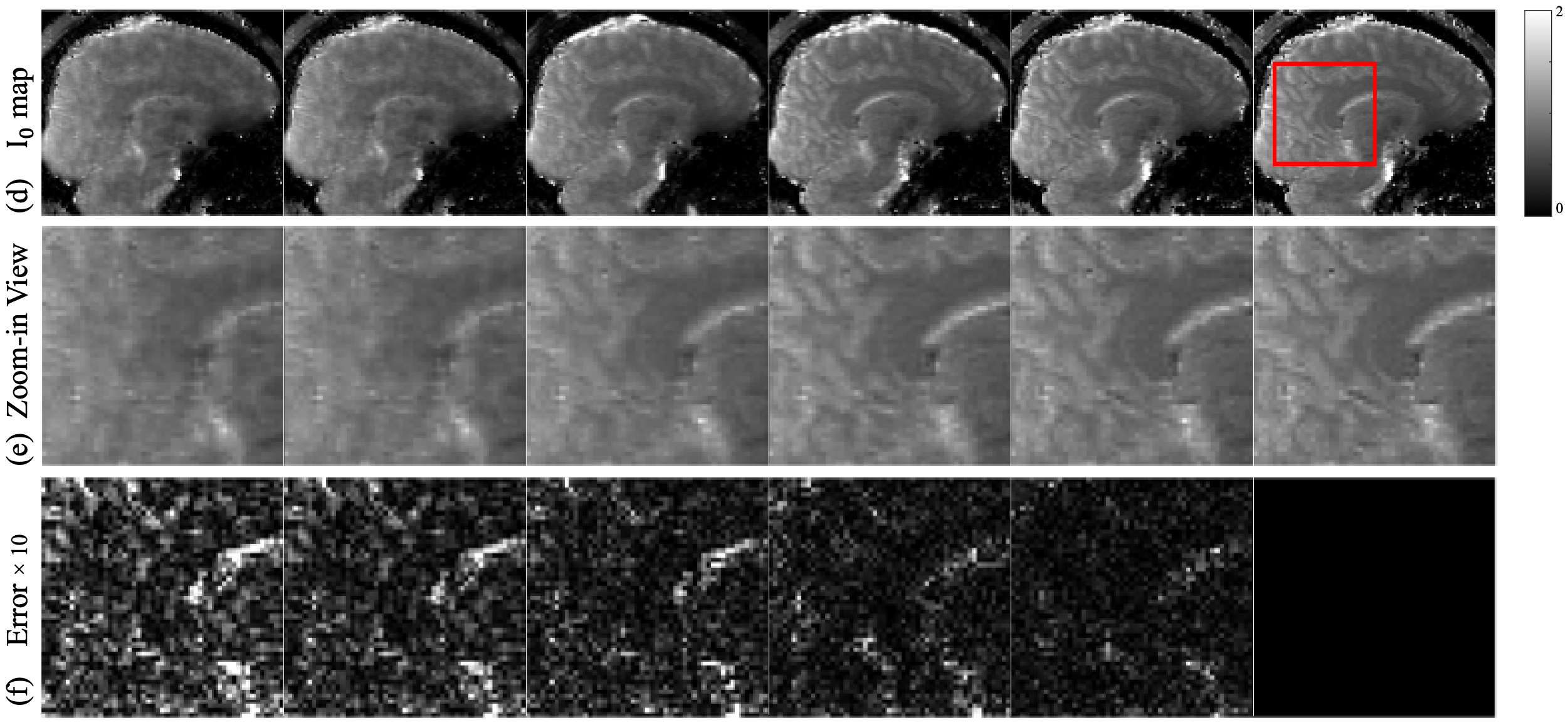}
\caption{Exemplified comparison of $T_1$  and $I_0$ mapping among different methods using 4$\times$ 2D Poisson undersampling
mask.}\label{fig:poisson}
\end{figure*} 
\begin{figure*}[htb]
\includegraphics[width=1\linewidth]{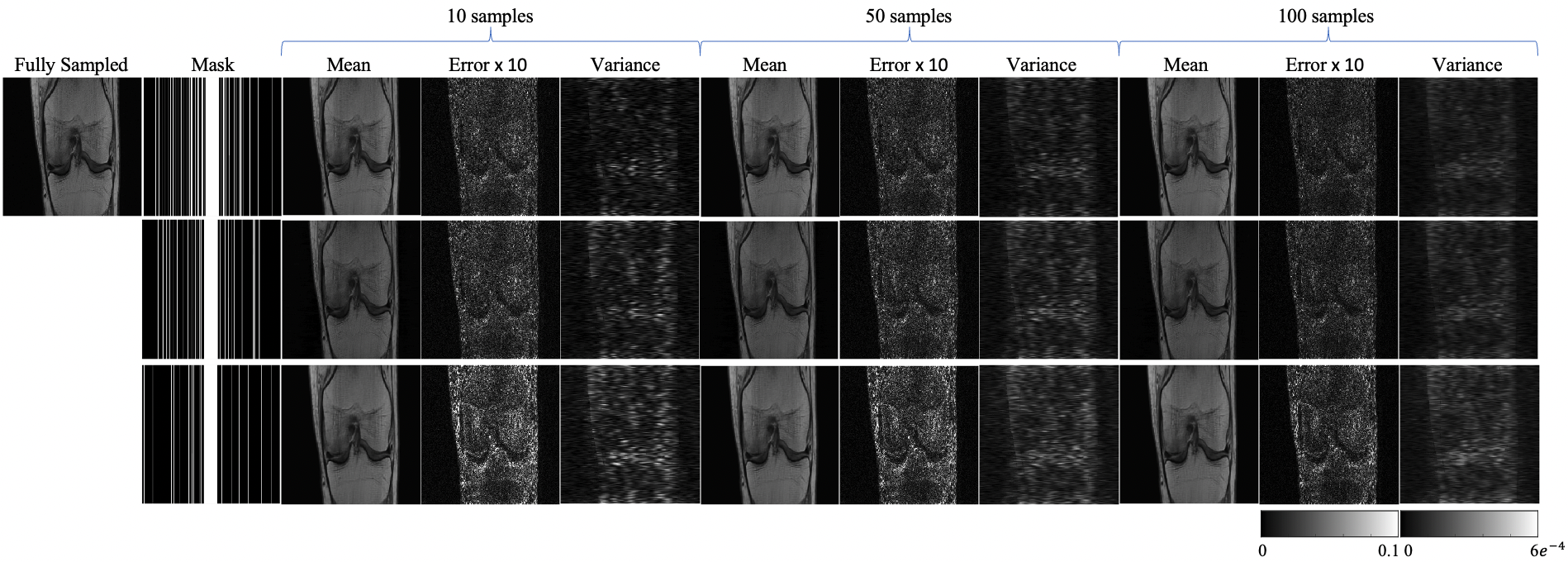}
\caption{Examples of mean, error and variance maps at different numbers of sampling for Static DiMo}\label{fig:uncertainty}
\end{figure*} 

The $I_0$ maps in Fig \ref{fig:cartesian}(d-f) exhibit a similar signature in reconstruction quality as the $T_1$ maps, where LLR and Model-TGV show aliasing artifacts with the most blur, further confirmed by the zoom-ins, reflecting the undersampling pattern of many high frequency data points being undersampled with low frequency data points fully sampled. The deep learning method RELAX  shows its ability to denoise through  artifacts removal and detail preservation. However, it still lacks sharpness resulting in smoothened edges.  Quantitative DiMo produces the sharpest maps preserving detail, which again is confirmed by its error maps, showing least error.

The $T_1$ and $I_0$ maps estimated from 4$\times$  2D Poisson disk undersampling are shown in Fig. \ref{fig:poisson}. Examining Fig. \ref{fig:poisson}(a) and Fig. \ref{fig:poisson}(d), it is apparent that the main differences compared to the 4x 1D undersampling case are the different artifact patterns, which in turn stems from the different undersampling patterns. As seen in the zoom-in views of the $T_1$ (Fig. \ref{fig:poisson}(b)) and $I_0$ (Fig. \ref{fig:poisson}(e)) maps, compared to Fig. \ref{fig:cartesian}, LLR provides sharper maps details but still retain artifacts and noise, whereas Model-TGV preserves detail but exhibits blurred edges due to its general averaging of noise. RELAX captures sharp structures but is not able to attain details in some fine regions. Quantitative DiMo produces artifact free $T_1$ and $I_0$ maps with superior performance compared to other methods. This is again confirmed by the error maps (Fig. \ref{fig:poisson}(c) and Fig. \ref{fig:poisson}(f)), with Quantitative DiMo showing the least error. This is likely achieved through integrating the unrolling gradient descent algorithm and diffusion denoising network, prioritizing noise suppression without compromising the fidelity and clarity of the underlying tissue structure.

\subsection{Ablation Study}
An ablation study investigating model uncertainty was conducted by sampling the well-trained Static DiMo on PD data 100 times. From these 100 samples, we derived mean images, error maps, and variance maps. We then compared three distinct acceleration factors (4$\times$, 5$\times$, 6$\times$) using 1D Cartesian masks, shown in Fig.\ref{fig:uncertainty}. When compared with fully sampled images, the error and variance maps predominantly capture the edge and boundary details. Notably, higher acceleration factors increases the uncertainty with amplified errors, especially around the tissue interfaces.
Incrementally increasing our sampling data counts, starting from 10, increasing to 50, and finally reaching 100, we witnessed a progression in the variance maps. Namely, they began displaying smoother transitions, with the high uncertainty edge regions gradually diminishing. This suggests that augmenting the number of sampling instances can be instrumental in attenuating noise within the mean images.

\section{Discussion and Conclusion}
While diffusion models, including ours, show impressive abilities on performing the image reconstruction task, there are still some limitations. For instance, diffusion process is computationally intensive, especially those involving complex structures or high-dimensional data, can be computationally expensive, requiring significant time and resources. Also, Diffusion models can be sensitive to the choice of parameters. A slight change in parameters can result in substantially different outcomes, making it crucial to fine-tune them for specific applications. For complex systems or datasets, it might be challenging to derive analytical solutions using diffusion models, necessitating the use of numerical methods that can introduce their own set of issues. Future research is warranted to explore to address these limitations.

To sum up, this paper proposed a diffusion model that is conditioned on native data domain for reconstructing MRI and quantitative MRI. The reconstruction shows promising results comparing to other deep learning methods, which reflects the robustness and efficiency of proposed Static DiMo and Quantitative DiMo.

\bibliographystyle{IEEEtran}
\bibliography{main}

\end{document}